# A Data-Theoretic Approach to Identifying Violent Facial Expressions in Social Crime Contexts

Arindam Kumar Paul
*Mathematics Discipline*
*Khulna University*
Khulna, Bangladesh
arindam017@gmail.com

*Abstract*— **Human Facial Expressions plays an important role for identifying human action or intention. Facial expressions can represent any specific action of any person and the pattern of violent behavior of any person strongly depends on the geographic region. Here we've designed an automated system by using a Convolutional Neural Network which can detect weather a person has any intention to commit any crime or not. Here we proposed a new method that can identify crime intensions or violent behavior of any person before executing crimes more efficiently by using a very little data of facial expressions before executing crime or any violent tasks. Instead of using image features which is a time-consuming and faulty method we used an automated feature selector CNN (Convolutional Neural Network) model which can capture exact facial expressions for training and then can predict that target facial expressions more accurately. Here we used only the facial data of a specific geographic region which can represent the violent and before crime facial patterns of the people of whole region.**

## I. INTRODUCTION

From the literature of violence detection, we can observe that no Machine learning or any other models are not being used by any industry or any individuals or any institution currently. But many scientists and processes are trying to build an Excel learning model for detecting violence since a long time ago. Actually final deduction is one of the most valuable and hardest tasks nowadays.

According to the Report of US ministry of Media Control and The influence of the cinema on children and adolescents, an annotated international bibliography by UNESCO media Actor's and highlighted Small Number of Negative Characters Influence on Human Psychology Media Actors influences the psychology, attitude, behaviors and other patterns of Childs and young population mostly. According to this study, in general the heroic characters are portrayed by movies and media as a Gangster especially they all are very efficient in committing violence. The fact is movies and media portray all of their works and attitudes positively. It means a negative violent attitude or behavior of the heroic character is justified by the portrayal of the director in the media. [2]

Media actors influences the psychology, attitude, behaviors and other patterns of Childs and young population.

Portrayal of minorities as a positive character influences the psychology attitude and behavior of children and youths mostly. The pattern of violence activities of a certain geographical region can be represented by their movie actors. A large portion of the total population and negatively affected by the justified and positively portrayed violence activities of the movie actors. Since the children and young people are mostly infected by this media and movies so they will reflect these types of patterns in their future life though. [3]

## II. PROPOSED METHODOLOGY

Since we can see that crime patterns as well as facial expressions and facial image properties varies very dynamically with changing situation, region, time, crime type and many more. Analyzing all of these is must needed in order to prevent the crime activities. As we know that, behavior and psychology of human beings of most of the regions or cultures are heavily influenced by their media and surroundings. Especially all the people of any certain region tries to follow the patterns they all see or observe on media and news. Almost all population of any region commonly try to copy the most common strategies they all are seeing on media and news. Media is covered by most of the population and the strategies they see on it influence their mind for executing any types of non-professional crimes in a same manner. Even sometimes the positive characters who is portrayed as racist, criminal or as a role model become the real role model for non-professional criminals. Their behavior, attitude, facial expressions, movements, strategies all are copied by maximum non-professional criminals at almost all crime activities. For these reasons the patterns of crimes become similar at a certain geographic area and it makes it easy to find out the crime activities [4].

We find out such some characters like media actors and they are:

- Media actors of a certain geographic region
- Negative Characters on television of a certain geographic region.
- Very Professional role model criminals of a certain geographic region.
- Criminal but public hero type characters of a certain geographic region.
- Such characters who become the news headline for their tactful criminal activities.
- Social media. activities of peoples of a certain geographic region.

If we analyze above mentioned character's facial expressions before they're executing the crime, some of the similar expressions should be found over the non-professional criminal's facial expressions before executing crimes at a same geographical region. But for gaining very accurate result from these types of data, analyzing more properties like body movement, voice, and characteristics should bring better result. [5]

## III. COLLECTING IMAGES

As we mentioned above, at first, we collected some pictures which contains all the properties of human faces who are attempting to commit a crime of those categories of people. These are our targeted pictures which we want to identify as they are attempting to do a crime. All of our collected photos are the faces of criminals just before the situation of doing a crime or planning for a crime. These images are used as samples of human faces before committing a crime. Next, we have collected some general facial photos seems normal human faces who are not suspected as criminals or have no intensions to commit any crimes of same geographical areas. These photos included smiley faces, talking or communicating people, working, playing, sitting, gaming and some other normal working time captures. All of these photos are collected from various websites and image search engines. Because it's quite hard to find an image database which contains the facial expressions of various peoples just before committing the crime, we selected the faces of various actors acting on a crime scene on various TV shows or movies. We selected the best matching photos of faces represents the pre-crime expressions perfectly on various situations. We collected the face images from various directions with different light intensity levels for our experiments. We included the faces of humans between age range 20-70 for both targeted sampling and negative sampling. We covered almost all categories of images for our experiment. We made two sets of images labeled as suspect and general face.

Figure 1 and 2 represents some of our training data of facial images from both positive and negative samples [6].

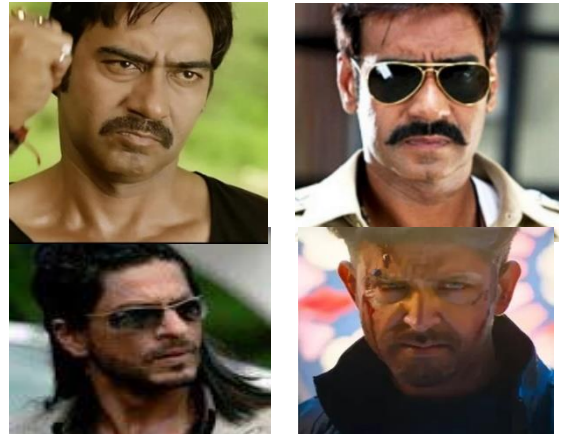

*Figure 1: Positive Samples*

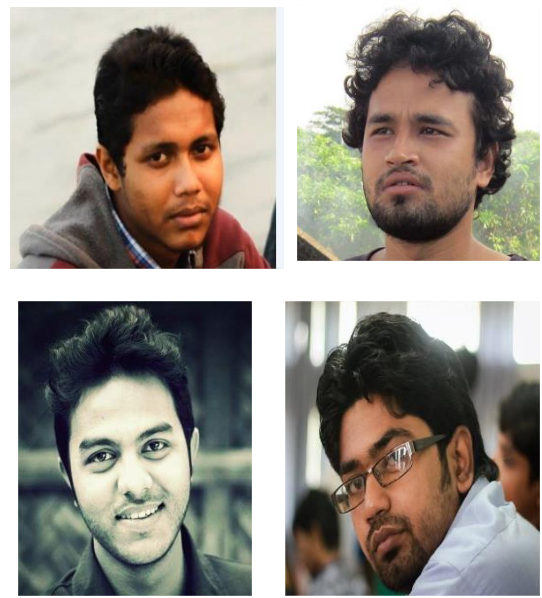

*Figure 2: Negative Samples*

## IV. DESCRIPTION OF MODEL AND PARAMETERS

We used a **Convolutional Neural Network** model for prediction purposes. We plotted the performance of convolutional neural network models with different loss functions. For the lack of computational power and high quality data within a short time we have used 650 images as training samples and 100 images as validation samples of the model.

Made prediction on known data and got a very accurate prediction in most of the cases.

We evaluated the performance of the model in terms of Accuracy and loss of the model. actually we tried to find out the optional model configuration such as optimizer, loss function, the number of hidden layers, fully connected layers, activation functions etc. that can model our small data set perfectly.

- A low accuracy and huge loss means huge errors on a lot of data.
- A low accuracy but low loss means little errors on a lot of data.
- Objective Situation: A great accuracy with low loss means low errors on a few data (best case).

- Current Situation: A great accuracy but a huge loss, means huge errors on a few data.

Since dataset is small and the images can be considered as not well balanced and noisy, we focused on the structure of the model capable of modelling award image data set and classify it well. All the images were resized to the dimension 100*100 pixels. Then we prepared our training and test data set by splitting it by the ratio 70% for training and 30% for validating. we have visualized our model structure below in Figure 3 [7].

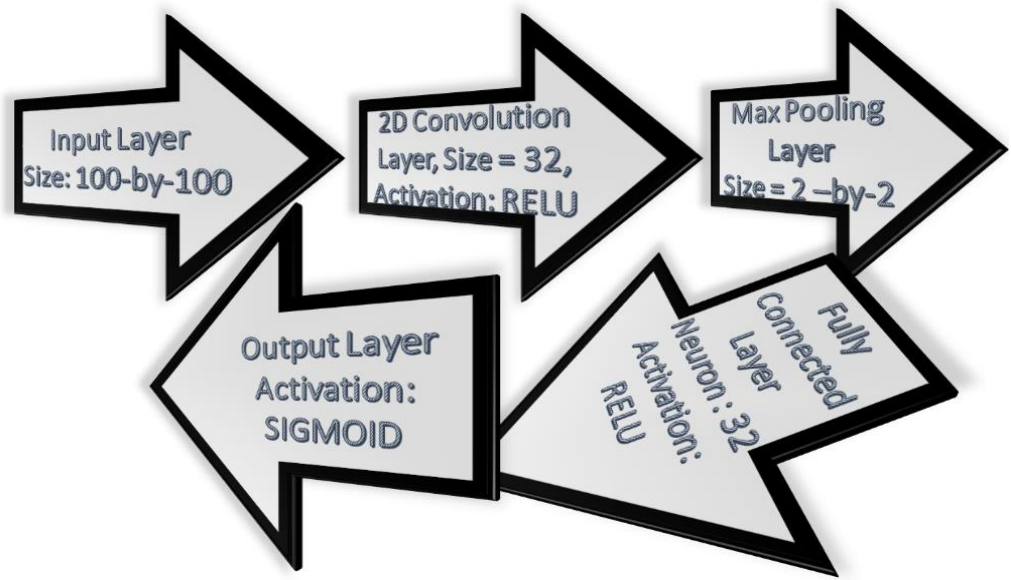

*Figure 3: CNN Model Architecture*

We have classified the images into two classes using a convolutional neural network but we have designed the prediction procedure in a different way. Instead of taking a class like violent and nonviolent, we calculated the probability of an individual being violent. This measurement gives us an accuracy of another dimension with which we can predict images more accurately. We have calculated all the outcomes and determine the outcome based on this probability. Firstly, we have set a scale for the Violent attitude. This can be considered as a situation all approach for solving this problem. Since our data is not cleaned and true accurate so we implemented this technique for gaining accurate results [8].

We resized images as 100×100 pixels before training. We've used the RELU and LeakyRELU as activation function for the 1st layer, RELU for fully connected layers and sigmoid for the output layer. We used ADAM and RMSPROP as optimizer functions and selected the final one by evaluating the performance. Binary cross entropy was used as loss function and evaluation metric was accuracy.

### A. Relu Activation Function

If x is the input vector, then we can define the RELU activation function as

$$R(x) = \begin{cases} x & x > 0 \\ 0 & x <= 0 \end{cases}$$

It avoids and rectifies vanishing gradient problem. ReLu is less computationally expensive than Tanh and sigmoid because it involves simpler mathematical operations.

### B. Adam Optimizer

Adaptive Moment Estimation (Adam) combines ideas from both RMSProp and Momentum. The Mathematical expression is:

$$V_{dW} = \beta_1 v_{dW} + (1 - \beta_1)\frac{\partial \mathcal{J}}{\partial W}, S_{dW} = \beta_2 s_{dW} + (1 - \beta_2)(\frac{\partial \mathcal{J}}{\partial W})^2$$

$$v_{dW}^{corrected} = \frac{v_{dW}}{1 - (\beta_1)^t}, s_{dW}^{corrected} = \frac{s_{dW}}{1 - (\beta_1)^t},$$

$$W = W - \alpha \frac{v_{dW}^{corrected}}{\sqrt{s_{dW}^{corrected}} + \varepsilon}$$

Where,

$V_{dW}$ = The exponentially weighted average of past gradients.

$S_{dW}$ = The exponentially weighted average of past squares of gradients.

β1, β2 = Hyper parameter to be tuned.

$\frac{\partial \mathcal{J}}{\partial W}$ = Cost gradient with respect to current layer.

W= The weight matrix (parameter to be updated).

$\alpha$ = The learning rate.

$\varepsilon$ = Very small value to avoid dividing by zero.

### C. Rmsprop Optimizer

Root Mean Square Prop (RMSProp) works by keeping an exponentially weighted average of the squares of past gradients. RMSProp then divides the learning rate by this average to speed up convergence.

$$s_{dW} = \beta s_{dW} + (1 - \beta)(\frac{\partial \mathcal{J}}{\partial W})^2$$

$$W = W - \alpha \frac{\frac{\partial \mathcal{J}}{\partial W}}{\sqrt{s_{dW}^{corrected}} + \varepsilon}$$

Where meaning of the symbols are same as of ADAM. [9]

## V. TRAINING AND TESTING THE MODEL

We collected images and then preprocessed them accordingly. After that we trained 4 CNN models with different configurations for achieving the best accuracy with minimum loss. We fitted the dataset to CNN and plotted the accuracy and loss value with respect to the epochs. By evaluating those figures, we selected the best model for our dataset and made prediction on new data.

The figures 4-11 is the accuracy and loss functional values of CNN model with different optimizers, activation, preprocessing and etc [10].

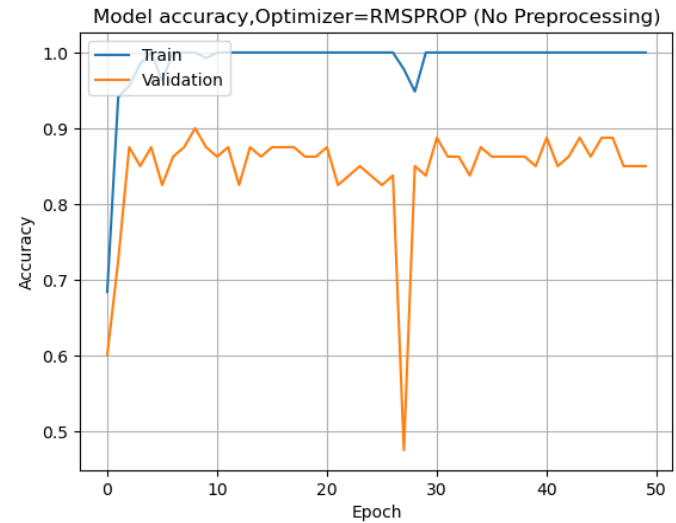

*Figure 4: Model accuracy with RMSPROP optimizer shows that the curve of training and validation accuracy have different patterns*

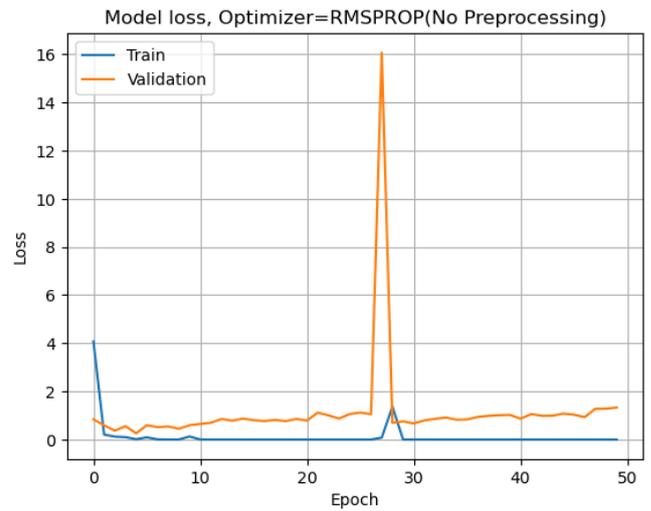

*Figure 5: Model loss with RMSPROP optimizer shows that the curve of training converges and validation doesn't.*

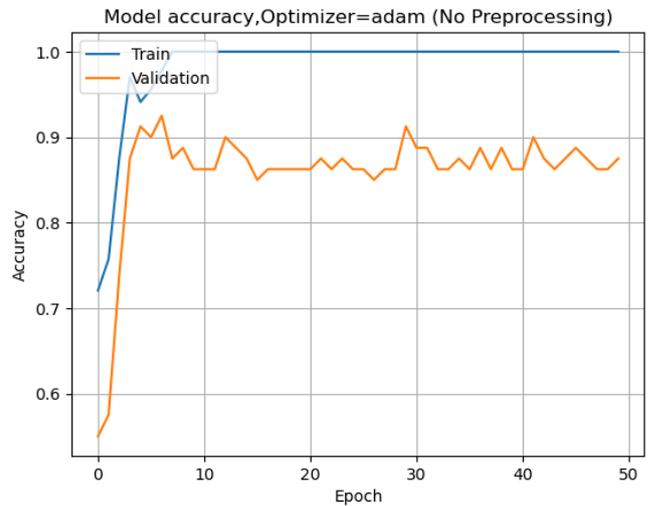

*Figure 6: Model accuracy with ADAM optimizer shows that the curve of training and validation accuracy have different patterns and validation accuracy indicates overfitting*

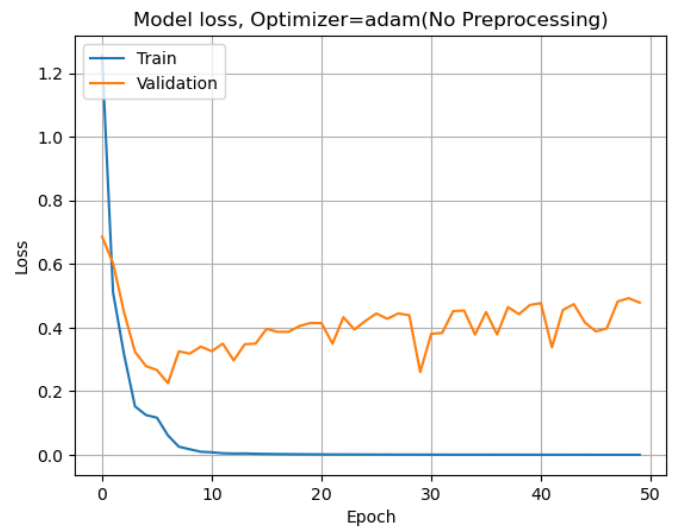

*Figure 7: Model loss with ADAM optimizer shows the validation loss curve is following non-convergent pattern*

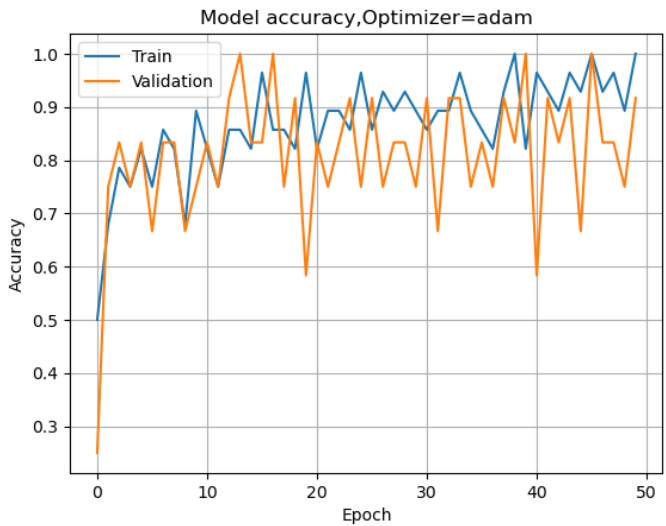

*Figure 8: Model Accuracy with ADAM and LeakyRelu shows a fluctuating but analogues pattern and training and validation curve is very close to each other during the whole time*

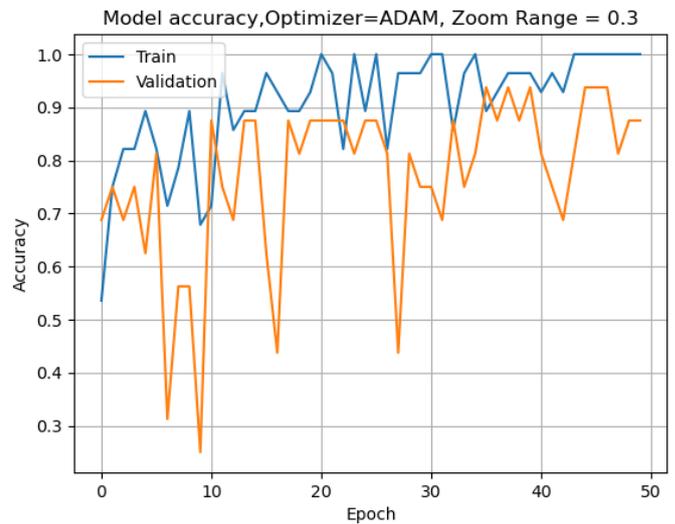

*Figure 10: Model Accuracy with random zooming images with ADAM optimizer shows training and validation accuracy curve is similar but preprocessing was not very accurate*

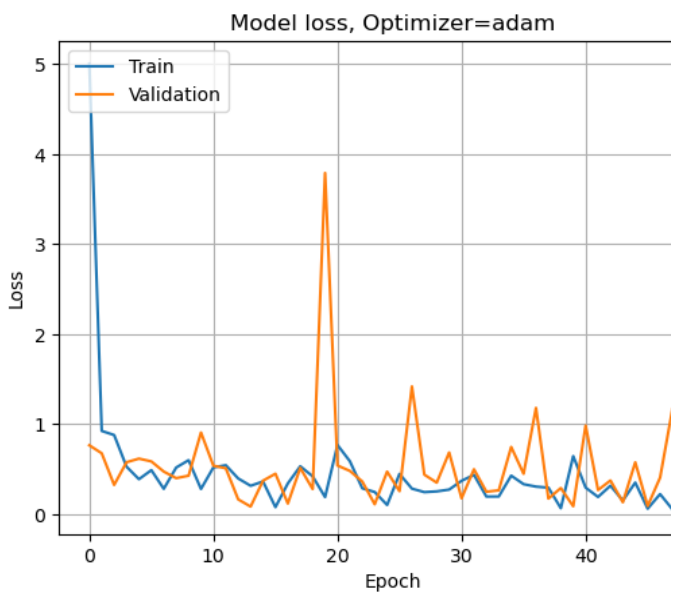

*Figure 9: Model loss with ADAM and LeakyRelu shows a monotonic behavior while decreasing and the pattern of training and validation loss is almost repeat the same pattern*

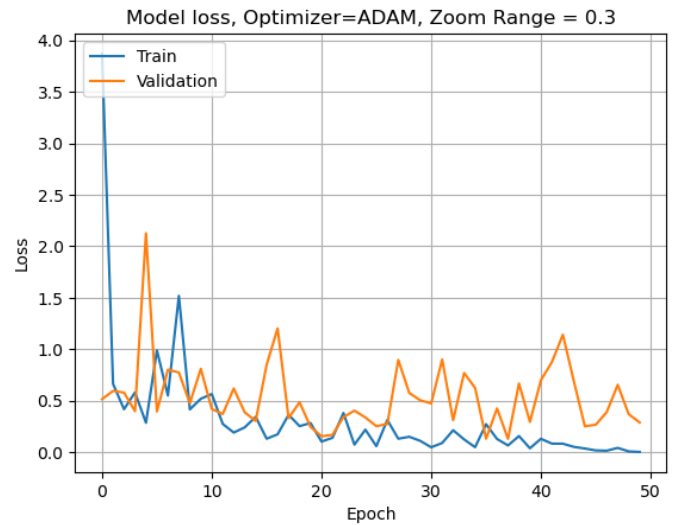

*Figure 11: Model loss with random zooming images with ADAM optimizer shows a convergence tenancy*

We trained 4 CNN models with our data set using different activation functions and optimizers. we used the LeakyRELU and RELU both in the hidden layers and Sigmoid for the output layer. We have trained models after doing some image processing and without any kind of processing. We have used to different optimizer and they are RMSPROP and ADAM. In the figure three we can see that the RMSPROP optimizer performed well in case of training. the accuracy was 100% and loss function converged well. but when we take a look at the validation curve then we can see that there is a difference between the training performance and validation performance and validation performance is not a satisfactory as the training performance. it indicates that the model is well balanced, the error rate is very low, the model was able to understand the underline pattern of the data but the model failed in case of predicting new data. The same thing goes for the combination of ADAM without any image preprocessing. ADAM was able to optimize the loss function video l while training the data set but it failed to perform accordingly in the case of new data. When we used images with a little preprocessing with zooming the images and then Training it using the optimizer ADAM, from the figure we can see that the training accuracy was fluctuating and the loss function value was also did not converged very well here but the performance of training and validation was almost the same in this case. in this case the training is higher and the error is also lowest. Specifically, the difference between the training performance and the validation performance I almost same. It indicates that, this model can predict new data successfully and the model is well balanced. In figure 7 and 8, we can see that, with a zooming, we get accuracy and loss value converges well and loss is the minimum with a 100% accuracy. So, in case of minimum or insufficient data, it's not possible to gain the accuracy at top and loss at minimum but using various processing, it's possible to optimize the performance of the model.

*Table 1: Model analysis*

| Model | Training Accuracy | Validation Accuracy | Training Loss | Validation Loss | Extra properties |
|---|---|---|---|---|---|
| *Model 1* | *100%* | *100%* | *0.0014* | *0.1058* | |
| **Model 2** | 99.26% | 86% | 0.0092 | 0.8600 | |
| **Model 3** | 100% | 86% | 0.000002` | 0.8253 | LeakyRelu in 1st layer |
| **Model 4** | 100% | 100% | $9.52 \times 10^{-4}$ | 0.0095 | Preprocessed image with a 0.2% zooming |

## VI. DISCUSSION

From the above figures and tables, it's clear that our proposed method is an effective and working method. We have predicted on new data with the trained model by an accuracy of 90%. So, there is no confusion that, in case of data unavailability, such methods can deal and produce the data successfully. For the better performance, more data should be collected and more preprocessing may provide better accuracy. We predicted new images using our trained CNN model and the outcome is shown in Figure 13. From Figure 13, it's clear that our model successfully captured the underlying pattern of the data and classified well. From the Figure 12 we can tell that, the facial expression should represent violent are predicted as violent and normal facial expressions are predicted as normal. We predicted total 48 images (21 violent, 27 normal) and 21 violent images was predicted as violent and 24 normal images was predicted as normal. Total 4 normal images were predicted as violent. Our model predicted 4 violent facial images as normal.

From Figure 12, the confusion matrix tells us the CNN model predicts well in case of both. It predicts the violent facial expressions very accurately and normal facial expressions too. Some of normal facial expressions was predicted as violent because the model counts all the features and sometimes in a same region some features are same for most of the human.

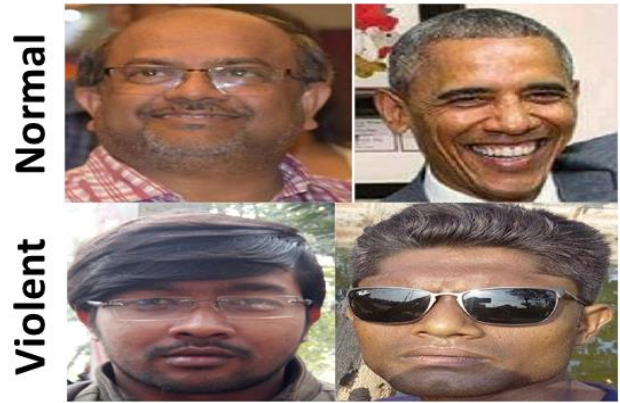

*Figure 13: Prediction on new images*

In Figure 13, we tested some images making some people acting as violent and others as normal. The model predicted all the violent acts and normal acts perfectly.

## VII. CONCLUSION

Here, we tried to develop a new strategy for solving with problems where the data is minimum or unavailable. We have connected the problem with a socio-behavioral theory to generate the dataset. We found our approach effective in practical. For the lack of computational power, we were unable to perform the experiment with sufficient data. More data should be gathered and processed wisely to make some real application from this method. We just modified the model to make the prediction more accurate but some data can solve the task easily. Image feature based approaches can give better outcome as well as making the process interpretable.

*Figure 12: Confusion Matrix*

**Biography**

Arindam Kumar Paul is an author, researcher, applied mathematician, and data science practitioner from Bangladesh. He completed his secondary education at Kumira Multipurpose Secondary School and his higher secondary education at Satkhira Government College.

He holds a B.Sc. in Mathematics and an M.Sc. in Applied Mathematics from the Department of Mathematics, Khulna University, Khulna, Bangladesh. He has been researching Machine Learning, Mathematical Modeling, Optimal Control, Deep Learning and Interpretability of ML/DL Models, System Theory & Complex Dynamical Systems, Uncertainty Modeling, Multiscale Dynamical Systems, Autonomous and Non-Autonomous Systems for more than 5 years. Currently he is designated as a Research Assistant at Khullna University. Arindam is a member of IEOM Society, Bangladesh Mathematical Biology Society. He reviews research articles and have collaborate with several researchers nationally and internationally.